\documentclass[acmsmall]{acmart}

\AtBeginDocument{%
  }

\setcopyright{acmlicensed}
\copyrightyear{}
\acmYear{}
\acmDOI{}

\acmISBN{}

\usepackage{amsmath,bm}
\usepackage{multirow, multicol}
\usepackage{rotating}
\usepackage{wrapfig}
\begin{document}


\title{Federated Learning for Smart Grid: A Survey on Applications and Potential Vulnerabilities}

\author{Zikai Zhang}
\email{zikaiz@unr.edu}
\affiliation{%
  \institution{University of Nevada, Reno}
  \city{Reno}
  \state{Nevada}
  \country{USA}
}

\author{Suman Rath}
\email{suman-rath@utulsa.edu}
\affiliation{%
  \institution{The University of Tulsa}
  \city{Tulsa}
  \state{Oklahoma}
  \country{USA}
}

\author{Jiahao Xu}
\email{jiahaox@unr.edu}
\affiliation{%
  \institution{University of Nevada, Reno}
  \city{Reno}
  \state{Nevada}
  \country{USA}
}

\author{Tingsong Xiao}
\email{xiaotingsong@ufl.edu}
\affiliation{%
  \institution{University of Florida}
  \city{Gainesville}
  \state{Florida}
  \country{USA}
}

\renewcommand{\shortauthors}{Zhang et al.}

\begin{abstract}
The Smart Grid (SG) is a critical energy infrastructure that collects real-time electricity usage data to forecast future energy demands using information and communication technologies (ICT). Due to growing concerns about data security and privacy in SGs, federated learning (FL) has emerged as a promising training framework. FL offers a balance between privacy, efficiency, and accuracy in SGs by enabling collaborative model training without sharing private data from IoT devices. In this survey, we thoroughly review recent advancements in designing FL-based SG systems across three stages: generation, transmission and distribution, and consumption. Additionally, we explore potential vulnerabilities that may arise when implementing FL in these stages. 
Furthermore, we discuss the gap between state-of-the-art (SOTA) FL research and its practical applications in SGs, and we propose future research directions. Unlike traditional surveys addressing security issues in centralized machine learning methods for SG systems, this survey is the first to specifically examine the applications and security concerns unique to FL-based SG systems. We also introduce FedGridShield, an open-source framework featuring implementations of SOTA attack and defense methods.
Our aim is to inspire further research into applications and improvements in the robustness of FL-based SG systems.
\end{abstract}



\maketitle

\section{Introduction}
The Smart Grid (SG), which integrates the physical infrastructure of the electrical grid with modern information and communication technologies (ICTs)~\cite{vasisht2022smart}, is a prime example of a cyber-physical system (CPS). Unlike conventional power grids, SGs can detect, respond to, and proactively manage changes in electricity demand and supply, while also addressing various operational challenges~\cite{obeagu2022overview}. 


The SG architecture incorporates several key components that work together to enhance the overall functionality of the electrical system~\cite{mittal2024smart}, including advanced metering infrastructure (AMI)~\cite{molokomme2019survey}, communication networks~\cite{hu2023multi}, and sophisticated information technology solutions~\cite{reza2014overview}, among others. By leveraging real-time data and advanced analytics, SGs aim to improve the efficiency of power generation, transmission and distribution, and consumption~\cite{chopda2021overview}. 
Although SGs enable a more flexible and resilient power system that can better accommodate the growing demand for electricity, 
the increasing reliance on ICTs has made SG systems more vulnerable to security and privacy attacks.
Federated Learning (FL) has emerged as a promising approach to addressing privacy and security concerns in SG systems. This innovative distributed machine learning paradigm enables collaborative model training across multiple participants without the need to share raw data, making it particularly suitable for SG applications, where generators are distributed and data privacy for each generator is crucial. We categorize the application of FL in SGs into three aspects: generation, transmission and distribution, and consumption.
From the perspective of generation~\cite{tuballa2016review} in SGs, FL is primarily being used to enhance renewable energy integration and forecasting~\cite{camacho2011control}. FL enables distributed SG generators to collaboratively train models for predicting resources such as solar and wind power output without sharing sensitive operational data. In the transmission~\cite{bayoumi2015power} and distribution aspects~\cite{hamidi2010smart} of SGs, FL is proving valuable for line fault detection and classification, optimal power flow~\cite{usman2022novel}, voltage control~\cite{xie2021review}, and more. By utilizing an FL-based approach, grid operators can quickly identify and locate faults in transmission lines, while collaboratively training models for power distribution without compromising the privacy of individual generators. In the consumption process of SGs~\cite{kaur2020energy}, FL is extensively used for load forecasting~\cite{husnoo2023secure} and energy theft detection~\cite{li2024fuse}. Smart meters can collaboratively train models to predict household energy consumption patterns without sharing individual usage data.

\begin{table}[htbp]
  \centering
  \caption{Compared with State-of-the-art Survey Papers.}
  \scalebox{0.93}{
    \begin{tabular}{cccccccc}
    \toprule[2pt]
    \textbf{Ref.} & \textbf{Year} & \textbf{Stage} & \textbf{Applications} & \textbf{DL-based} & \textbf{Privacy} & \textbf{Security} & \textbf{FL-based} \\
    \midrule
    Grataloup et al.~\cite{grataloup2024review} & 2024  &       & $\surd$ & $\surd$ & $\surd$ &       & $\surd$ \\
    Zheng et al.~\cite{zheng2024advancing} & 2024  &       & $\surd$ & $\surd$ & $\surd$ &       & $\surd$ \\
    Zhang et al.~\cite{zhang2024vulnerability} & 2024  & $\surd$ & $\surd$ & $\surd$ & $\surd$ & $\surd$ &  \\
    Pandya et al.~\cite{pandya2023federated} & 2023  &       & $\surd$      & $\surd$ & $\surd$ &       & $\surd$ \\
    Cheng et al.~\cite{cheng2022review} & 2022  &       & $\surd$ & $\surd$ & $\surd$ &       & $\surd$ \\
    Mirzaee et al.~\cite{mirzaee2022smart} & 2022  &       & $\surd$ &       & $\surd$ & $\surd$ &  \\
    Ding et al.~\cite{ding2022cyber} & 2022  & $\surd$ & $\surd$ &       & $\surd$ & $\surd$ &  \\
    Cui et al.~\cite{cui2020detecting} & 2020  &       & $\surd$ & $\surd$ & $\surd$ & $\surd$      &  \\
    \midrule
    \textbf{Ours} & -  & $\surd$ & $\surd$ & $\surd$ & $\surd$ & $\surd$ & $\surd$ \\
    \bottomrule[2pt]
    \end{tabular}%
    }
  \label{tab:survey}%
\end{table}%

\subsection{Related Surveys}\label{subsec:related_surveys}
We summarize the state-of-the-art (SOTA) survey papers~\cite{grataloup2024review,zheng2024advancing,zhang2024vulnerability,pandya2023federated,cheng2022review,mirzaee2022smart,ding2022cyber,cui2020detecting,gao2025flowertune} focus on SGs, with particular attention to seven aspects: whether the paper discusses the stages of SGs, compares SG applications, includes Deep Learning (DL)-based methods, discusses privacy and security issues, and whether it focuses on FL-based SG systems (see Table~\ref{tab:survey}).

{While existing surveys have explored various facets of FL or SG security, a dedicated and comprehensive analysis of FL-based SG systems—encompassing diverse applications alongside their specific security and privacy vulnerabilities—remains less developed. For instance, works by Grataloup et al.~\cite{grataloup2024review} and Pandya et al.~\cite{pandya2023federated} touch upon FL applications in renewable energy and broader smart city contexts respectively, yet provide limited detail on SG-specific FL security issues. Other surveys, such as Zheng et al.~\cite{zheng2024advancing}, detail FL components without addressing model-related attacks within FL-based SG systems, and Cheng et al.~\cite{cheng2022review} offer a concise review of FL for SGs but lack depth on robustness. Conversely, while Zhang et al.~\cite{zhang2024vulnerability}, Mirzaee et al.~\cite{mirzaee2022smart}, Ding et al.~\cite{ding2022cyber}, and Cui et al.~\cite{cui2020detecting} provide valuable insights into general SG security, common privacy violations, and specific threats like False Data Injection Attacks, their primary focus often remains on traditional centralized machine learning paradigms rather than the unique challenges introduced by FL. Our paper aims to bridge this crucial gap by providing a focused, comprehensive discussion on FL-based SGs. We systematically examine their applications across different SG stages and conduct a detailed analysis of their potential security and privacy vulnerabilities, thereby complementing existing research and intending to inspire further specialized studies at this critical intersection.}

\subsection{Key Contributions}

\begin{enumerate}
\item We provide a systematic study of existing research on FL-based SG applications, privacy issues, and security concerns. By defining and categorizing these into three stages (i.e., generation, transmission and distribution, and consumption) of SG applications, and two categories (i.e., privacy and security) of vulnerabilities, our holistic survey offers a hierarchical understanding of FL's usage in SGs to preserve data privacy.
\item We compare our paper with other SOTA survey papers across seven aspects: whether the paper discusses the stages of SG applications, compares SG applications, includes DL-based methods, discusses privacy and security issues, and focuses on FL-based SG systems. This highlights the focus of our survey, which centers on FL-based SG applications and their potential vulnerabilities.
\item We discuss the use cases of the FL framework in three stages of SG systems. For the generation stage, we cover tasks such as energy forecasting, energy optimization, and energy management. In the transmission and distribution stage, FL has been applied to tasks such as energy distribution, attack detection, anomaly detection, and security assessment. In the consumption stage, FL can enhance tasks like optimal power usage, load forecasting, and energy theft detection. Incorporating FL at these stages has improved data privacy and distribution efficiency in SGs. After introducing FL applications to these stages, we discuss the potential vulnerabilities for FL-based SGs in both privacy and security contexts.
\item {Our survey extensively discusses SOTA attack and defense methods in FL and identifies a significant gap between current research and its practical application in FL-based SG systems. To help bridge this gap and empower researchers to explore the SOTA strategies discussed, we introduce FedGridShield. This open-source framework provides implementations of these critical attack and defense methods, serving as a practical tool for pursuing the future research directions we further suggest in this domain.}
\end{enumerate}

\subsection{Structure of the Paper}

\begin{figure}[h]
\includegraphics[width=14cm]{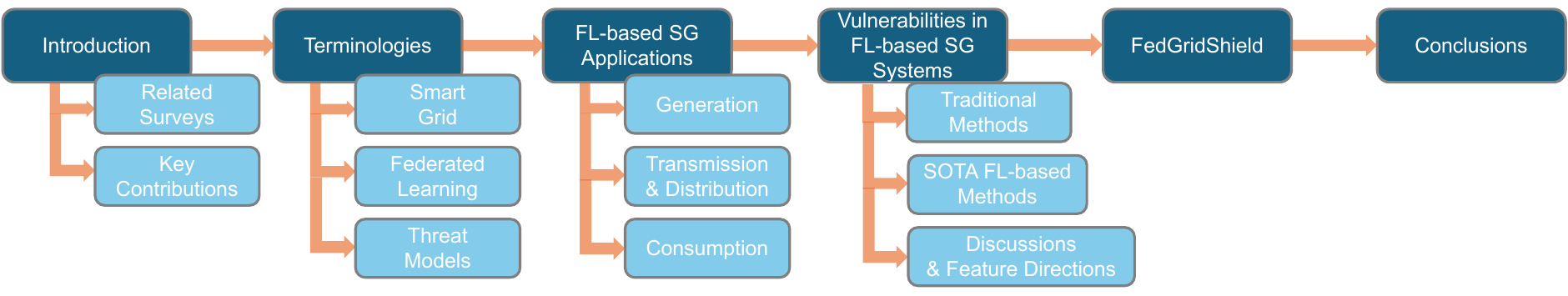}
\caption{{Structure of the Paper.}
\vspace{-8pt}
\label{fig:structure}}
\end{figure}

The structure of our paper is illustrated in Figure~\ref{fig:structure}. In the following sections, we begin by exploring the foundational terminologies of SGs, the FL framework, and its threat models in Section~\ref{sec:terminology}. Next, Section~\ref{sec:application} provides a comprehensive review of FL-based SG applications, detailing their implementation across various stages. 
{In Section~\ref{sec:vulnerabilities}, we thoroughly examine vulnerabilities within SG applications. This analysis covers traditional adversarial methods as well as SOTA FL-based attack and defense methods relevant to these systems. Within this section, we also identify existing research gaps and propose future directions in the field. Subsequently, in Section~\ref{sec:fedgridshield}, we introduce our FedGridShield framework and detail its implementation.}
Finally, Section~\ref{conclusion} concludes the paper.


\section{Terminologies}\label{sec:terminology}

\subsection{Smart Grid}
The electric power grid has evolved by leaps and bounds since its inception \cite{masera2018smart}. A major development in this domain was the integration of ICT. The integration of ICT with the traditional power grid facilitated distributed power generation via better resource management and automated control \cite{simoes2012comparison}. ICT integration also marked the evolution of the conventional grid into the SG and solved several limitations, including unidirectional power flow and centralized bulk generation of electricity. Additionally, it reduced the power and energy domain's dependence on fossil fuels like coal and natural gas, leading the way for sustainability and decarbonization via the development of microgrids \cite{jahromi2023cyber, yadav2020microgrid}. 
The SG power flow framework consists of the following stages:
\begin{enumerate}
    \item Generation: This stage refers to the production of AC/DC power. In the conventional power grid, power generation was centralized in nature. However, with ICT integration for real-time monitoring, power generation can also be achieved in a regulated manner at residential buildings (for example, using solar photovoltaic panels) and used for local consumption/fed back to the grid \cite{kahrobaee2012multiagent}.
    \item Transmission/Distribution: This stage refers to the transfer of the power produced at the generation stage to the appropriate end nodes for utilization. This involves the use of transformers, transmission lines, substation equipment, and other line devices like fault detector relays, breakers to ensure a safe and reliable transfer at appropriate bus voltage levels. The fault detector relays and breakers are often digital in nature and controlled by the central monitoring station via ICT devices \cite{gaspar2023smart}.
    \item Consumption: This is the final stage of the SG power flow process and involves the utilization of the power as received from the transmission and distribution networks for powering electrical devices. The consumption level can involve entities like government agencies, private industries, private individuals, etc., each of which can have a different consumption footprint. In the modern SG, integration of ICT devices means that some consumer-level devices like electric vehicles can also be used to transfer surplus/unused power back to the grid (coordinated via V2G communications \cite{madawala2011bidirectional}).
\end{enumerate}

While ICT integration in the power grid enabled easier and more robust resource management from a single centralized Supervisory Control and Data Acquisition (SCADA)-based control system \cite{gaushell1993scada}, the increasing number of IoT devices and local generation facilities like solar panels means that it can be extremely computationally resource-intensive to monitor them. Inadequate monitoring facilities can mean that intrusive events like cyberattacks and line faults may not be detected and mitigated on time.

Slowed mitigation of cyberattacks, faults, and other control emergencies can have a devastating impact on the stability of the SG \cite{jahromi2023cyber}.
Moreover, the SG being a cyber-physical system implies that cyber-layer vulnerabilities can be exploited to cause severe physical ramifications. For example, cyberattacks on the SG can lead to uncontrolled voltage increase at one or more buses leading to equipment damage and consumer-level device burnout \cite{rath2022behind}. To alleviate these issues, several researchers have proposed, implemented, and evaluated the use of Artificial Intelligence (AI)-based tools and techniques for applications like real-time control \cite{bose2017artificial}, cybersecurity \cite{ruan2023super}, load forecasting \cite{zhou2017holographic}, and fault detection \cite{sun2023high}.
These AI tools are typically deployed at the centralized controller level for monitoring, removing the need for human system operators.

As the AI modules employed for applications like cybersecurity/control monitoring and load forecasting in the SG environment, are often placed at a central server, they require the exchange of data between local Distributed Energy Resources (DERs) and the central node via communication networks.
However, this presents privacy risks \cite{sun2021solar} as the data being transferred can be utilized to reveal energy consumption patterns of private individuals \cite{liu2018practical}.
This is a major risk as the consumption patterns can be cross-analyzed to determine when the individuals are at home/outside. Further, their home appliances can also be revealed motivating malicious entities to schedule burglary attempts \cite{liu2018practical}.

Another risk of the communication-based training data exchange between local DERs and the centralized server housing AI modules is the non-willingness within third-party DER owners to share proprietary data.
This non-willingness typically stems from the concern that the proprietary data may be accessed by malicious entities who can then know the system details of the DER.
This can be used to manipulate the system via stealth cyberattack vectors \cite{leng2021stability}.
Non-willingness can also arise from a lack of trust between the DER owners and the utility company that manages the SG and the control center.

Both the above-mentioned factors can lead to data constraints that may severely hinder the AI training process, leading to flawed and untrustworthy decision-making, raising concerns on the reliability of AI.
To solve the problem, an AI training strategy called FL can be used. In this strategy, proprietary training data does not need to be transferred over a communication channel, leading to alleviation of privacy concerns \cite{mcmahan2017communication}. More details about this training strategy are available in the following subsection.




\subsection{Federated Learning}\label{subsec:FL}

FL~\cite{zhang2024fed,zhang2025fed} aims to collaboratively train models across multiple decentralized data owners holding potentially sensitive local data in a privacy-preserving manner. The essence of FL lies in its ability to learn a shared model by aggregating locally computed updates, rather than directly accessing or sharing the raw data. This machine learning (ML) paradigm not only enhances privacy and security but also enables the utilization of distributively owned data for model training, making it particularly suitable for mission-critical applications, such as healthcare and finance, where data privacy is of paramount concern.

The typical FL process involves several key steps. Initially, a global model is distributed to all participating FL clients from the FL server. Each client then trains the ML model on its local data to derive an updated local model. These model updates are subsequently sent back to the FL server, where they are aggregated to update the global model. This cycle is repeated until convergence or specific performance criteria are met. A widely used framework in FL is Federated Averaging (FedAvg), renowned for its ability to aggregate model updates with minimal communication overhead. This is particularly crucial in FL settings, where a potentially large number of participants may have limited communication bandwidth.

To achieve significant real-world impact, FL must overcome several complex challenges. These include the prevalence of non-independently and identically distributed (non-IID) data across clients, systemic heterogeneity, and scalability issues. Non-IID data can introduce biases, favoring the data distributions of particular participants. System heterogeneity, arising from variations in computational power and communication capabilities among clients, can lead to uneven contributions to the model training process. Furthermore, as the number of participants grows, scalability becomes a concern, necessitating efficient algorithms and robust infrastructure for managing update aggregation and global model distribution. The field of FL is continuously evolving, driven by advancements in both theoretical research and practical applications across diverse industries, and is anticipated to play a pivotal role in developing AI solutions that prioritize user privacy and data sovereignty.

Consider a typical FL setup involving $m$ clients (e.g., IoT devices such as smart grid sensors) and a central server (e.g., a smart grid server). These clients collaborate to train a global model $\mathbf{w} \in \mathbb{R}^p$, with dimension $p$, coordinated by the central server. Assume each client $j \in [m]$ possesses a local private dataset $\mathcal{D}_j$. The primary objective of FL is to solve the following optimization problem:
\begin{equation} \label{eq:fed_obj}
\min_{\mathbf{w} \in \mathbb{R}^p} F(\mathbf{w}) := \frac{1}{m} \sum_{j \in [m]} F_j(\mathbf{w})
\end{equation}
where $F_j(\mathbf{w}) := \mathbb{E}_{z \in \mathcal{D}_j} [\ell(\mathbf{w}; z)]$ represents the local loss function for client $j$, and $\ell(\mathbf{w}; z)$ is the loss function dependent on the model parameter $\mathbf{w}$ and a data point $z$ sampled from $\mathcal{D}_j$. It is important to note that the datasets $\mathcal{D}_j$ and $\mathcal{D}_k$ may exhibit different distributions for $j \neq k$.

{To solve this optimization problem, the system operates for $T$ rounds of FL training. Initially, the server stores the global model $\mathbf{w}^0$. In each round $t \in [T]$, the server randomly selects a subset $\mathcal{Q}^t$ of $q$ clients and disseminates the latest global model $\mathbf{w}^t$ to them. These selected clients then update the global model $\mathbf{w}^t$ using their respective local data. For instance, in the widely-used FedAvg algorithm, each selected client $j \in \mathcal{Q}^t$ performs $\kappa$ iterations of stochastic gradient descent (SGD) to update its local model as follows:
\begin{equation} \label{eqn:local-sgd}
\mathbf{w}_j^{t,r+1} = \mathbf{w}_j^{t,r} - \alpha \cdot \mathbf{g}_j^{t,r}, \quad \forall r = 0, \ldots, \kappa - 1
\end{equation}
where $\mathbf{w}_j^{t,r}$ denotes the local model of client $j$ at local iteration $r$ of round $t$, initialized as $\mathbf{w}_j^{t,0} = \mathbf{w}^{t}$. Here, $\alpha$ is the local learning rate, and $\mathbf{g}_j^{t,r} := \frac{1}{B} \sum_{z \in \xi_j^{t,r}} \nabla \ell(\mathbf{w}_j^{t,r}, z)$ is the stochastic gradient computed over a mini-batch $\xi_j^{t,r}$ of size $B$ sampled from $\mathcal{D}_j$. After local training, each client computes the model update $\Delta_j^t := \mathbf{w}^{t} - \mathbf{w}_j^{t,\kappa}$ and sends it to the server. In round $t$, the server aggregates these updates and applies them to the global model. However, the presence of malicious actors necessitates considering potential attacks.}



\subsection{Threat Models}\label{subsec:threat_models}

We consider different types of adversaries who can compromise the integrity and privacy of the FL system. Let $\mathcal{A}$ denote a generic attacker.

{\textbf{Privacy Inference Attackers.} Let $\mathcal{A}_{priv}$ denote a privacy inference attacker. The {objective} of $\mathcal{A}_{priv}$ is to deduce sensitive information about clients' local data $\mathcal{D}_j$ by observing the shared model updates $\Delta_j^t$ or the aggregated global model $\mathbf{w}^t$. Their capabilities ($\mathcal{C}_{priv}$) includes,
\begin{itemize}
    \item Possess diverse prior information (e.g., auxiliary datasets, knowledge of model architecture) and computational resources.
    \item May collude with other malicious entities (clients or an adversarial server).
    \item Typically do not inject fabricated data with the intent of degrading the model, but passively observe communications to infer private data.
    \item Can be an ``honest-but-curious'' server or participating clients.
\end{itemize}}


{\textbf{Byzantine Attackers.} Let $\mathcal{A}_{byz}$ denote a Byzantine attacker. These attackers aim to disrupt the learning process or degrade the performance of the global model. The {objective} of a general Byzantine attacker, $\mathcal{A}_{byz}$, within the FL system is to undermine the overall performance of the trained global model $\mathbf{w}$ as much as possible. This is also known as an \textit{untargeted model poisoning attack}. Let $k_{byz}$ be the number of Byzantine attackers among the $m$ clients, where typically $k_{byz} < m/2$ for many defense strategies to be effective. Their capabilities ($\mathcal{C}_{byz}$) often includes:
\begin{itemize}
    \item Maliciously alter their local SGD processes (Equation~\eqref{eqn:local-sgd}), for example, by flipping gradients, adding large noise to updates, or submitting random model weights.
    \item May generate and submit arbitrary model updates $\tilde{\Delta}_{byz}^t$ that do not correspond to legitimate training on their local data.
    \item May collude with other Byzantine attackers to coordinate their malicious updates.
    \item Due to secure aggregation, they typically remain unaware of the model updates $\Delta_j^t$ of benign clients in the same round before aggregation.
\end{itemize}}

{\textbf{Backdoor Attackers.} Let $\mathcal{A}_{bd}$ denote a backdoor attacker. This is a specific type of model poisoning attacker with a more targeted malicious goal. The objective of $\mathcal{A}_{bd}$ is to implant a hidden backdoor into the global model $\mathbf{w}$. The backdoored model $\mathbf{w}_{bd}$ should behave normally on standard inputs from the primary task distribution but produce specific, attacker-chosen incorrect outputs when inputs contain a predefined trigger pattern (e.g., a small pixel patch in an image, a specific phrase in text).
Their capabilities ($\mathcal{C}_{bd}$) includes:
\begin{itemize}
    \item Control a fraction of the participating clients (let this be $k_{bd}$ clients).
    \item Modify the training data $\mathcal{D}_j$ on their controlled clients by injecting samples that pair the backdoor trigger with the target malicious label.
    \item Manipulate the local model training process (Equation~\eqref{eqn:local-sgd}) on their controlled clients to generate model updates $\Delta_j^{t,bd}$ that steer the global model towards learning the backdoor task.
    \item Aim for the backdoor to be persistent and effective in the aggregated global model $\mathbf{w}$ without significantly degrading its performance on the main task for benign inputs (to evade detection).
    \item May possess knowledge of the model architecture and the aggregation scheme.
\end{itemize}}

\section{FL-based SG Applications}\label{sec:application}

\begin{figure}[h]
\includegraphics[width=12cm]{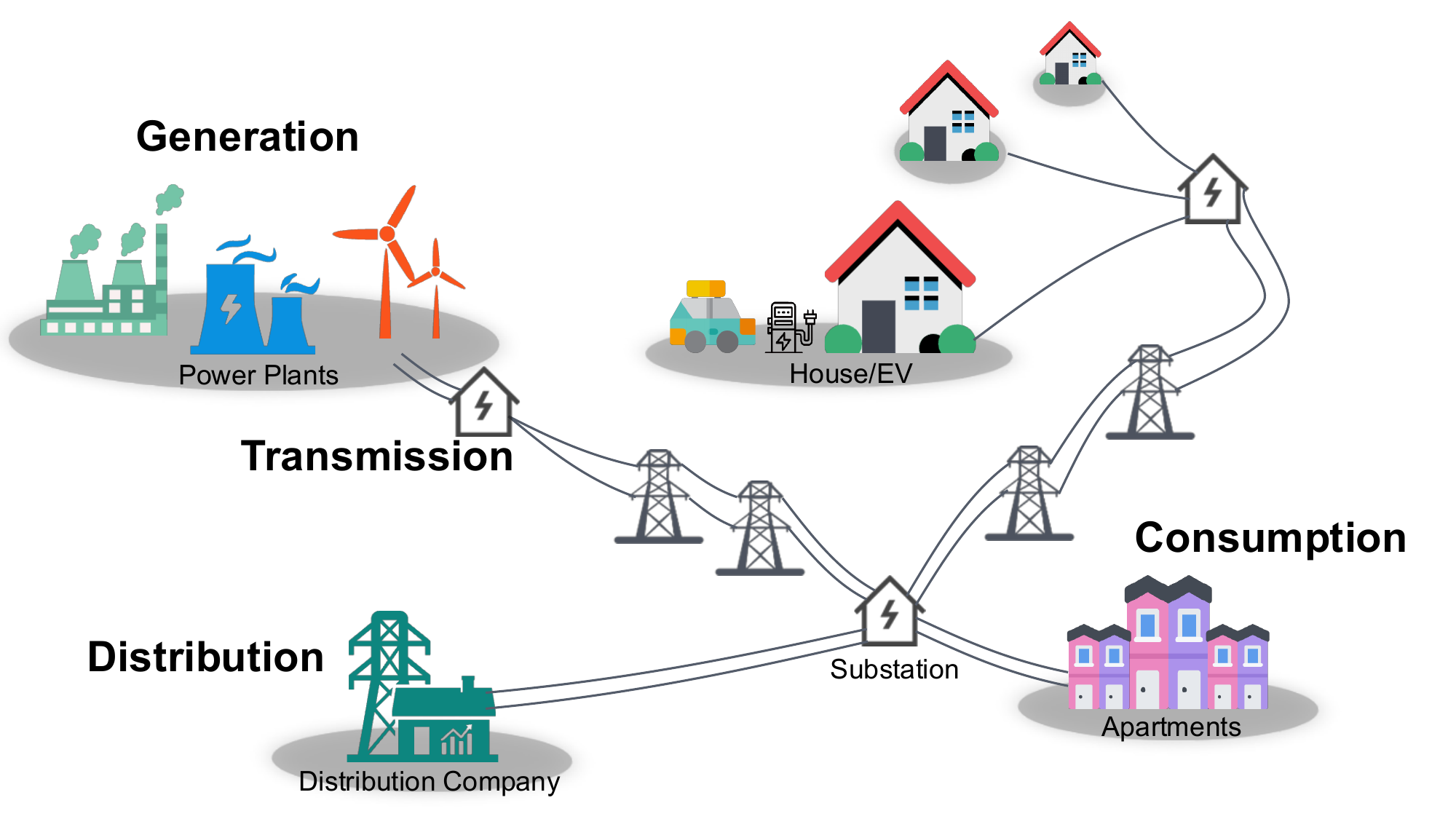}
\caption{Stages of SG Systems. Electricity is produced by utilities and independent power producers at various types of power plants. The electric transmission system serves as a crucial connection between power generation and consumption. High-voltage transmission lines transport electricity across long distances from power plants to populated areas. At substations, the voltage is stepped down to lower levels, enabling distribution companies to deliver the power to local communities. Finally, the electricity is used by consumers for everyday purposes at the point of consumption.\label{fig:applications}}
\vspace{-8pt}
\end{figure}

The process of delivering electricity to consumers involves several key stages (as shown in Figure~\ref{fig:applications}). First, electricity is generated at power plants by utilities and independent power producers using various energy sources such as fossil fuels, wind, and solar. Next, the transmission system carries the electricity over long distances through high-voltage lines from the generation sites to substations. At substations, the voltage is reduced to lower levels for safe distribution. The distribution network then delivers this electricity to homes, businesses, and other consumers. Finally, the electricity is used by consumers for daily activities, powering appliances, devices, and industrial equipment. 
\begin{wrapfigure}{r}{7cm}
\includegraphics[width=7cm]{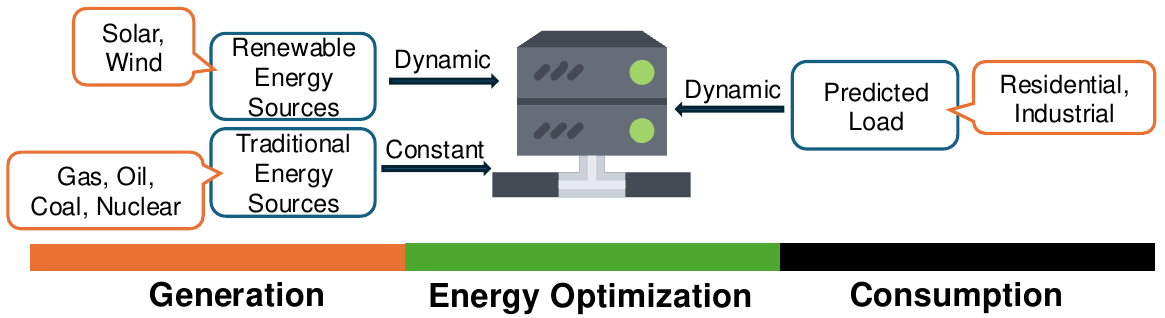}
\caption{The importance of energy optimization in complex resource and consumption scenarios.\label{fig:3_1_energy_optimization}}
\end{wrapfigure}
FL has emerged as a powerful framework for various SG applications, offering privacy-preserving and distributed solutions for tasks such as energy forecasting, energy optimization, energy management, energy distribution, attack detection, anomaly detection, security assessment, optimal power usage, load forecasting, and energy theft detection, among others.
In the following subsections, we discuss how each of these works utilizes FL in SG applications.

\subsection{Generation Stage}

FL enables collaborative model training across different energy producers and consumers, enhancing the accuracy of energy generation forecasts and enabling dynamic resource allocation without the need to centralize sensitive operational data~\cite{lee2023towards, gupta2023fedgrid}. In the context of energy management~\cite{abdellatif2024sdcl}, FL supports the development of intelligent systems that can predict and optimize power generation patterns while keeping sensitive user data local on smart devices. This approach improves grid efficiency and facilitates more effective demand response programs.


\textbf{Energy Forecasting.}
In the work by \cite{lee2023towards}, an advanced algorithm for FL is proposed to tackle the challenges of non-IID data in distributed solar energy grids. This framework is specifically designed to handle the heterogeneous nature of the data across different solar energy producers, enhancing the model's ability to converge efficiently in decentralized settings. By addressing the variations in data distribution, the framework improves the accuracy and reliability of predictions, leading to better performance in solar energy generation and more effective resource allocation.

\textbf{Energy Optimization.}
The FedGrid framework introduced by \cite{gupta2023fedgrid} offers a secure FL solution for optimizing energy management within smart grids. This framework emphasizes the importance of protecting privacy and enhancing data security during the optimization process. By enabling efficient collaboration between various grid components without compromising data confidentiality, FedGrid helps optimize energy distribution and usage. The approach also addresses the key challenges associated with data sharing and collaboration in smart grids, ultimately improving the overall efficiency and reliability of the system. As shown in Figure~\ref{fig:3_1_energy_optimization}, energy optimization is crucial for coordinating multiple energy resources and consumption, especially when using FL techniques to protect data privacy.

\textbf{Energy Management.}
In \cite{abdellatif2024sdcl}, a Secure, Distributed, and Collaborative Learning (SDCL) framework is presented to enhance the operations of smart grids. This framework focuses on maintaining data privacy and security while facilitating collaborative learning among grid components. By incorporating advanced AI and blockchain technologies, SDCL aims to optimize grid performance, improve system efficiency, and ensure reliable energy management. Additionally, \cite{lin2024privacy} introduces a privacy-preserving energy management framework that utilizes vertical FL for smart data cleaning in smart homes. This approach enhances user privacy while enabling effective analysis and management of smart meter data, ensuring that energy usage is optimized without compromising sensitive information.

\subsection{Transmission and Distribution Stages}
For energy distribution~\cite{mourtzis2023integration}, FL enables optimal energy distribution solutions and flexibility predictions, significantly improving the efficiency of power distribution in SGs. 
In the area of attack~\cite{tran2023efficient, zhao2024federated} and anomaly detection~\cite{kesici2024detection, jithish2023distributed}, FL has demonstrated its effectiveness in identifying cyber attacks and anomalies within SG systems. This approach allows multiple stakeholders to collaboratively train robust detection models while maintaining data privacy, as raw consumption data remains local. 
In the realm of assessment~\cite{li2024privacy, ren2023qfdsa, ren2023efeddsa}, FL has been applied to dynamic security assessment, enabling participants to collaboratively build models while preserving data privacy.

\textbf{Energy Distribution.}
Mourtzis et al.~\cite{mourtzis2023integration} introduce an FL framework designed to enhance the efficiency and security of energy distribution within smart grids, while concurrently addressing privacy concerns by enabling collaborative model training without the exchange of raw data. A key feature of their proposed system is an incentive mechanism where service providers compensate edge devices based on their contributions to designated tasks.

This mechanism is structured around two primary payoff functions. For an edge device, the payoff function balances the revenue earned—influenced by factors such as the timeliness or "Age of Local Model" contributed—against its operational costs, which encompass local data processing, computation, and communication expenses. For a service provider, the corresponding payoff function weighs the accuracy achieved by the aggregated global model against the total payments made to participating edge devices and the costs incurred for cloud-based aggregation. The objective of this incentive structure is to encourage robust participation and high-quality contributions from edge devices while allowing service providers to obtain accurate models in a cost-effective manner.

\textbf{Attack Detection.}
Tran et al.~\cite{tran2023efficient} introduce a privacy-preserving FL approach specifically designed to detect false data injection attacks (FDIAs) in smart grid systems. This method focuses on enhancing data privacy and security in cross-silo FL scenarios within the SG environment. By improving the detection efficiency and effectiveness of FDIAs, their approach ensures that data confidentiality is maintained across various entities involved in the grid, making it a robust solution for securing SG infrastructure.
Zhao et al.~\cite{zhao2024federated} propose a FL algorithm that integrates client sampling and gradient projection techniques to enhance SG applications. This approach effectively addresses privacy concerns by enabling model training using fragmented data from devices spread across the grid. Furthermore, the inclusion of client sampling and gradient projection techniques leads to improved FL performance and optimization, providing a more reliable framework for handling SG data without compromising security.

\textbf{Anomaly Detection.}
Kesici et al.~\cite{kesici2024detection} propose a data-driven time series anomaly detection method that leverages time-aware shapelets, a shapelet evolution graph, and segment embeddings learned through the DeepWalk algorithm. Their method demonstrates a significant improvement in detecting FDIA, with a 20-40\% increase in accuracy compared to other unsupervised learning techniques. Additionally, when applied with BO-XGBoost, the model achieves a 5\% higher accuracy than unmodified XGBoost, adding further robustness to the detection process. Moreover, this approach imparts physical significance to the dynamic evolution of time series models, providing valuable insights into the underlying data.
Jithish et al.~\cite{jithish2023distributed} present an FL-based approach for distributed anomaly detection in SGs. Their solution enhances privacy and security by allowing for collaborative detection of anomalies without the need to centralize sensitive data. This distributed approach not only safeguards data privacy but also improves the overall stability and reliability of SG systems. By enabling effective anomaly detection across the grid, their method ensures that potential issues are identified early, contributing to more resilient SG operations.

\textbf{Security Assessment.}
Several approaches aim to enhance security and privacy in FL for SG dynamic security assessment (DSA). Li et al.~\cite{li2024privacy} introduced QFDSA, a system that integrates FL with quantum-resistant cryptography to bolster data privacy and security against advanced threats, including those from quantum computing. Building on this, Ren et al.~\cite{ren2023qfdsa} further developed QFDSA, emphasizing the use of Quantum Key Distribution (QKD) protocols, such as Measurement-Device-Independent QKD (MDI-QKD). In these QKD systems, participants (like ``Alice'' and ``Bob'') exchange quantum states, often via a relay ( ``Charlie''), to establish highly secure cryptographic keys. The effectiveness and security of these keys, crucial for protecting the FL process, are determined by analyzing factors like signal transmission probabilities and error rates, without needing to delve into the complex underlying mathematical formulas for a general understanding.

In a distinct but related effort, Ren et al.~\cite{ren2023efeddsa} proposed EFedDSA, an efficient horizontal FL approach that incorporates differential privacy for SG dynamic security assessment. This method strengthens privacy by protecting individual data contributions during the collaborative security analysis, thereby addressing data isolation challenges and improving the overall efficiency and confidentiality of DSA in SG environments.

\subsection{Consumption Stage.}
%
In the area of optimal power usage~\cite{rajesh2024federated, abdulla2024smart}, FL facilitates the extraction of typical electricity consumption patterns from smart meter data, enabling more efficient grid management.
For load forecasting~\cite{badr2023privacy, liu2023fedforecast, fekri2023asynchronous}, FL-based methods have been developed to enhance the accuracy of short-term residential load predictions by utilizing distributed data from multiple sources.
In the case of electricity theft detection~\cite{li2024fuse, alshehri2024deep, zafar2023step}, frameworks like FedDetect employ FL to effectively identify fraudulent activities while preserving consumer privacy.

\textbf{Optimal Power Usage.}
Rajesh et al.~\cite{rajesh2024federated} propose a FL framework designed to enhance privacy while optimizing power usage in SGs. The framework allows for collaborative learning across multiple participants without compromising individual user data. By securing power traces and enabling personalized recommendations for power usage optimization, the approach significantly improves privacy preservation in SG data analytics. This personalized aspect ensures that recommendations are tailored to the specific needs of users while maintaining data confidentiality.
Similarly, Abdulla et al.~\cite{abdulla2024smart} present an adaptive FL approach for energy consumption forecasting using smart meter data. Their method improves forecasting accuracy by utilizing smart meter data without sharing raw information, ensuring user privacy. The adaptive nature of the approach allows it to adjust to individual consumption patterns, providing personalized forecasting models for each participant. This enhances both the accuracy of energy predictions and the privacy of users in the SG system.

\textbf{Load Forecasting.}
Badr et al.~\cite{badr2023privacy} introduce a privacy-preserving and communication-efficient energy prediction scheme based on FL for SGs. Their approach enables accurate load forecasting while maintaining data confidentiality, which is crucial for infrastructure planning and power dispatching. By enhancing communication efficiency in the FL process, this method helps reduce power outages and equipment failures, making it an important tool for SG operations.
Liu et al.~\cite{liu2023fedforecast} propose FedForecast, a FL framework designed for short-term probabilistic individual load forecasting. This framework provides comprehensive information about future load uncertainties while addressing data isolation and privacy concerns inherent in traditional centralized forecasting methods. By maintaining data confidentiality across households or smart meters, FedForecast improves the accuracy and reliability of short-term load forecasts.
Fekri et al.~\cite{fekri2023asynchronous} develop an asynchronous adaptive FL algorithm for distributed load forecasting with smart meter data. This approach enhances the scalability and efficiency of load forecasting in SGs by adapting asynchronously to the data, thus addressing issues of privacy, security, and communication overhead found in traditional centralized methods. By doing so, the algorithm ensures accurate load forecasting while preserving the privacy of distributed data.

\textbf{Energy Theft Detection.}
Li et al.~\cite{li2024fuse} introduce FUSE, a framework that combines FL with U-shape split learning for detecting electricity theft. FUSE identifies abnormal electricity consumption behaviors while preserving the privacy of both user data and the detection model. By integrating FL with split learning, FUSE enhances data protection and model security in smart grid systems, offering a robust solution for electricity theft detection.
Alshehri et al.~\cite{alshehri2024deep} develop a deep anomaly detection framework that utilizes FL to detect electricity theft, including zero-day threats, in smart power grids. This framework preserves privacy while effectively identifying new and previously unseen attacks, thereby enhancing the security and stability of SGs. By addressing the challenge of detecting sophisticated electricity theft attempts, the framework plays a critical role in maintaining grid reliability.
Zafar et al.~\cite{zafar2023step} present a FL-assisted hybrid deep learning model for electricity theft detection using smart meter data. This model enhances the security and reliability of SGs, particularly in the context of Industry 5.0. By leveraging FL to keep data decentralized, the model addresses privacy concerns while effectively detecting electricity theft. This hybrid approach offers a comprehensive solution to safeguard SGs from fraudulent activities.


\section{Vulnerabilities in FL-based SG Systems}\label{sec:vulnerabilities}

\begin{figure}[t]
\includegraphics[width=11cm]{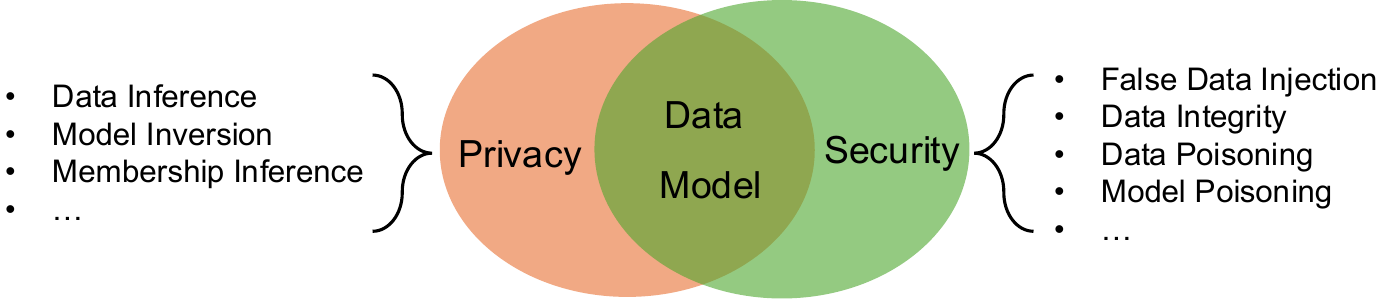}
\caption{The classification of vulnerabilities in FL-based SG systems.\label{fig:vulnerability}}
\vspace{-13pt}
\end{figure}

In this section, we explore the vulnerabilities present in SG systems. As illustrated in Table~\ref{tab:vulnerbility}, our discussion is structured around three critical stages within SGs: generation, transmission and distribution, and consumption, as well as two vulnerability aspects, as depicted in Figure~\ref{fig:vulnerability}: privacy and security. This systematic approach allows us to comprehensively demonstrate the potential vulnerabilities associated with various applications within these stages. In each subsection, we begin by presenting the fundamental concepts underlying each application to establish a clear understanding of their functionalities and objectives. We then examine the specific aspects that make these applications potentially vulnerable, highlighting their inherent risks.


\subsection{Traditional Adversarial Attack and Defense Methods}

\subsubsection{Vulnerabilities in Generation Stage}

In the context of SG, generation refers to the production of electricity from various sources, including both traditional and renewable energy~\cite{tuballa2016review}. It is a crucial component of the SG pipeline, serving as the starting point for the flow of electricity throughout the entire system.
As generation systems become more integrated and digitalized, privacy and security become increasingly important~\cite{krause2021cybersecurity}. Protecting generation infrastructure from attacks is essential for maintaining the reliability and integrity of the entire SG system. 

\textbf{Energy Forecasting.} 
Accurate energy forecasting is vital for integrating intermittent renewable energy sources into SGs~\cite{wang2023investigation}, often leveraging machine learning techniques on historical and meteorological data~\cite{malik2022integration}. However, this reliance on data-driven models, whether centralized or distributed, introduces vulnerabilities to adversarial manipulation.
FL approaches for energy forecasting attempt to address these issues, often by adapting defenses against threats similar to those seen in traditional systems. For instance, Shabbir et al.~\cite{shabbir2024resilience} explore how False Data Injection Attacks (FDIAs)—a common concern in centralized grid monitoring—can impact FL-based energy forecasting. Their work shows that various FDIA manifestations, such as scaling or random noise attacks on local data, can significantly degrade the FL model's predictive accuracy. Addressing fundamental data privacy concerns, which are critical in any data-handling system, Badr et al.~\cite{badr2022privacy} propose a privacy-preserving FL approach for net-energy forecasting. They employ techniques like secure data aggregation and differential privacy to protect against inference attacks on shared model parameters; these defense strategies are also foundational in protecting sensitive data in many centralized machine learning applications. Similarly, the FedGrid framework~\cite{gupta2023fedgrid} uses FL for renewable energy and load forecasting, inherently aiming to mitigate the privacy risks associated with traditional centralized data aggregation, where pooling sensitive user consumption and weather data would make it a prime target for conventional data breaches. FedGrid enables collaborative training without direct raw data sharing, thus addressing a key vulnerability of centralized architectures.
\begin{table}[t]
  \centering
    \caption{Paper Reviews on Vulnerabilities in SG Systems.\label{table:vulnerablity}}%
  \scalebox{0.55}{
    \begin{tabular}{c|cccccc}
    \toprule[2pt]
    \textbf{Stage} & \textbf{Ref} & \textbf{Year} & \textbf{Task} & \textbf{Data Type} & \textbf{Attack} & \textbf{Attack Type} \\
    \midrule
    \multirow{3}[2]{*}{\begin{sideways}Gen\end{sideways}} 
          & Shabbir et al.~\cite{shabbir2024resilience} & 2024  & Energy Forecasting & Time Series & False Data Injection & Security \\
          & Badr et al.~\cite{badr2022privacy} & 2022  & Net-Energy Forecasting & Time Series & Model Inversion, Membership Inference & Privacy \\
          & FedGrid~\cite{gupta2023fedgrid} & 2023  & Net-Energy Forecasting & Miscellaneous & Data Inference & Privacy \\
    \midrule
    \multirow{4}[2]{*}{\begin{sideways}T\&D\end{sideways}} 
          & Lin et al.~\cite{lin2022incentive} & 2022  & State Estimation & Measurements & False Data Injection & Security \\
          & Mohammadabadi et al.~\cite{mohammadabadi2023towards} & 2023  & Event Classification & Phasor Measurement Unit & Data Inference & Privacy \\
          & Jithish et al.~\cite{jithish2023distributed} & 2023  & Anomaly Detection & Miscellaneous & Data Integrity & Security \\
          & Liu et al.~\cite{liu2022federated} & 2022  & Voltage Control & Miscellaneous & Data Inference & Privacy \\
    \midrule
    \multirow{5}[2]{*}{\begin{sideways}Consumption\end{sideways}} 
          & Husnoo et al.~\cite{husnoo2023secure} & 2023  & Load Forecasting & Historical & Data Integrity & Security \\
          & Wen et al.~\cite{wen2021feddetect} & 2021  & Energy Theft Detection & Irish Smart Meter & Inference, Model Poisoning & Privacy, Security \\
          & Ashraf et al.~\cite{ashraf2022feddp} & 2022  & Energy Theft Detection & Smart Meter & Data Poisoning & Security \\
          & Zafar et al.~\cite{zafar2023step} & 2023  & Energy Theft Detection & Irish Smart Meter & Data Poisoning & Security \\
          & Li et al.~\cite{li2024fuse} & 2024  & Energy Theft Detection & Smart Meter & Data Poisoning, Model Inversion & Privacy, Security \\
    \bottomrule[2pt]
    \end{tabular}%
    }
  \label{tab:vulnerbility}%
  \vspace{-13pt}
\end{table}%

\subsubsection{Vulnerabilities in Transmission and Distribution Stages}

SG technology revolutionizes traditional power grids by incorporating advanced ICTs into the existing energy infrastructure. This integration enables a more efficient, reliable, and flexible system for electricity transmission and distribution~\cite{hamidi2010smart}. The transmission component of the SG~\cite{bayoumi2015power} focuses on the high-voltage transport of electricity over long distances. In SGs, transmission systems benefit from advanced technologies such as Flexible AC Transmission Systems (FACTS) and High-Voltage Direct Current (HVDC) power transmission, which enhance the controllability, stability, and power transfer capability of AC transmission systems. However, transmission systems are vulnerable to various types of attacks. FDIAs are particularly concerning, as they can manipulate sensor measurements and mislead grid operators.
SG distribution networks~\cite{konstantelos2016strategic} are responsible for delivering electricity to end-users at lower voltages. These networks are undergoing significant changes to accommodate the increasing integration of DERs and renewable energy sources. Smart distribution systems enable bi-directional power flow, allowing consumers to also become producers of electricity. However, distribution systems face threats such as energy theft through meter tampering, privacy breaches of consumer data, and coordinated attacks on demand response systems~\cite{ebtia2024spatial}. 

\textbf{State Estimation.}
AC State Estimation is a crucial process in SG systems, designed to determine the most likely steady-state operation of the power system based on field measurements. It serves as a fundamental data processing tool responsible for supporting and increasing system visibility while filtering errors that may appear in real-time measurements, system topology, and parameters~\cite{acurio2023state}. 

However, despite its significance, AC State Estimation is vulnerable to cyber threats, particularly FDIAs. These attacks can manipulate measurement data, introduce errors in state estimation, which can impact bus power flow calculations and the economic dispatch of the power system~\cite{huang2022false}.
Within the FL framework, Lin et al.~\cite{lin2022incentive} propose an edge-based FL framework for detecting FDIAs in SG state estimation. This approach leverages local system operating data to predict and assess system stability in a distributed manner. By utilizing edge devices for local model training, the framework reduces the computational burden on central servers and enhances privacy protection. A key innovation in this work is the incorporation of an incentive mechanism designed to encourage high-quality data sharing and model training. This mechanism addresses the challenge of motivating participants to actively contribute to the FL process, which is essential for improving the overall detection accuracy of the system.

\textbf{Events Classification and Anomaly Detection.}
Grid event classification and anomaly detection are vital for maintaining SG reliability. Deep learning models are often used but are known to be vulnerable to traditional adversarial attacks, where attackers make subtle changes to input data to cause misclassifications, a common concern even in centralized systems~\cite{niazazari2020attack}.

Addressing these issues, Mohammadabadi et al.~\cite{mohammadabadi2023towards} investigated the impact of such adversarial attacks (e.g., using the Fast Gradient Sign Method (FGSM)) on their proposed federated XGBoost model for event classification with Phasor Measurement Unit (PMU) data. FGSM and similar techniques craft malicious inputs by adding small, calculated perturbations to clean data to deceive machine learning models. While their framework uses FL for privacy and homomorphic encryption to secure communications, their consideration of FGSM directly engages with how these established attack methods affect distributed learning systems. They also explored data processing techniques as a defense. Separately, Jithish et al.~\cite{jithish2023distributed} utilized FL for real-time anomaly detection in smart meters, focusing on privacy. To protect the integrity and confidentiality of model updates during transmission—a fundamental security concern in any distributed network—they employed standard secure communication protocols like SSL/TLS.

\textbf{Voltage Control.}
Voltage regulation is a critical process in electrical power systems, designed to maintain voltage levels within acceptable limits across the network. It involves controlling and adjusting the voltage to ensure stable and reliable power delivery to consumers~\cite{xie2021review}. However, these networks are susceptible to data-related attacks, which can lead to voltage instability. Attackers may exploit vulnerabilities in communication channels to inject false data or disrupt the flow of information critical for effective voltage control. 

In the FL-based SG system, Liu et al.~\cite{liu2022federated} combine FL with reinforcement learning techniques to develop a privacy-preserving and efficient control strategy. In this federated reinforcement learning framework, grid-connected interface inverters of renewable energy sources are modeled as deep neural network-based agents that interact with their local environment to learn optimal voltage-var control strategies. The process operates in a two-stage process: local training using reinforcement learning algorithms, followed by periodic global aggregation of model updates. This approach offers several advantages, including enhanced data privacy, reduced communication costs, improved scalability, and adaptability to changing network conditions.

\subsubsection{Vulnerabilities in Consumption Stage}

The consumption part of an SG represents the end-user side of the energy distribution system~\cite{kaur2020energy}, encompassing homes, businesses, industries, and more. It integrates advanced technologies like AMI with smart meters for real-time usage monitoring and demand-side management to optimize energy consumption patterns. This system leverages IoT devices and real-time analytics to enhance energy efficiency and provide consumers with detailed insights into their usage. Additionally, sophisticated forecasting techniques and machine learning algorithms are employed to predict and manage energy loads effectively, ensuring a balance between supply and demand. However, smart meters and IoT devices, which are integral to these systems, can be vulnerable to attacks~\cite{liu2021vulnerability}, potentially allowing malicious customers to manipulate energy consumption data or disrupt power supply.

\textbf{Load Forecasting.}
Smart meter measurements, while essential for accurate demand forecasting, raise significant privacy and data breach concerns for consumers. The fine-grained load data collected by smart meters can potentially reveal sensitive information about residential users' daily routines and behaviors, posing risks to property security~\cite{husnoo2023secure}.
In the FL framework, Husnoo et al.~\cite{husnoo2023secure} propose a secure FL approach designed for residential short-term load forecasting. This approach addresses the critical balance between accurate demand prediction and protecting consumer privacy in smart meter data. FL allows collaborative model training without exposing raw data, as models are trained locally, and only parameters are shared with a central server. To enhance security, the framework incorporates advanced techniques like gradient quantization using SignSGD~\cite{bernstein2018signsgd} and differential privacy~\cite{dwork2006differential}. Additionally, it employs hierarchical and peer-to-peer strategies to improve both privacy and accuracy. The system tackles challenges such as non-IID data distribution and potential attacks through methods like deep federated adaptation and secure aggregation. Experimental results demonstrate that this approach achieves high forecasting accuracy while preserving user data privacy, with significant improvements in prediction accuracy compared to traditional methods.

\textbf{Energy Theft Detection.}
Energy theft detection~\cite{salinas2015privacy} in SGs refers to the process of identifying and preventing unauthorized consumption of electricity, an illegal activity that causes significant economic losses to utility companies. SGs integrate digital technologies with traditional power systems, enabling advanced energy management and monitoring capabilities. These systems utilize AMI and smart meters to collect, measure, and analyze energy usage data from customers at regular intervals. However, the collection and analysis of consumer energy data create vulnerabilities, potentially allowing attackers to compromise smart meters.


Within the FL framework, Wen et al.~\cite{wen2021feddetect} introduce FedDetect, a novel privacy-preserving FL framework designed for energy theft detection in SGs. FedDetect leverages FL to train machine learning models on distributed devices, aggregating local model updates instead of raw data to protect user privacy.
Ashraf et al.~\cite{ashraf2022feddp} propose FedDP, a privacy-protecting theft detection scheme for SGs using FL. This framework addresses the challenge of utilizing consumer energy data for theft detection while maintaining privacy protection.
Zafar et al.~\cite{zafar2023step} develop a FL-assisted hybrid deep learning model for electricity theft detection in SGs. Their model combines FL techniques with deep learning to enhance security and reliability of electricity theft detection.
Li et al.~\cite{li2024fuse} present FUSE, a novel electricity theft detection framework that integrates FL with U-shape split learning. FUSE aims to address the challenges of data privacy and security in SG systems while effectively detecting electricity theft.

\subsection{SOTA FL-based Attack and Defense Methods}\label{discussion}

In the previous sections, we summarize the applications of FL in SGs and the potential vulnerabilities discussed in the existing literature. Most works still focus on attack methods commonly used in centralized machine learning frameworks, such as FDIAs, untargeted adversarial attacks, and inference attacks. However, we propose that, compared to the SOTA research, there remains a significant gap between FL-based attack methods and the attack strategies explored in FL-based SG applications. In the following subsection, we will introduce SOTA FL attacks (i.e., Byzantine attacks, backdoor attacks, and inference attacks) and SOTA FL defense mechanisms. Finally, we will discuss several potential future research directions.

\subsubsection{SOTA FL Attack}\label{subsubsec:sota_fl_attack}

{\textbf{(1) Byzantine Attack in FL.} 
Byzantine attacks in FL involve malicious clients sending incorrect or misleading model updates to the central server, aiming to disrupt the training process and degrade model performance~\cite{fang2020local}. These malicious local model updates are typically intentionally modified by clients controlled by an attacker. We can categorize several classical and commonly discussed Byzantine attacks based on their approach to manipulating updates.}
\unskip

{\textbf{Naive Attacks.} In their simplest form, naive attack methods aim to disrupt the convergence of the global model without sophisticated manipulation. One such approach is the \textit{Random Attack}, where malicious clients send randomized updates drawn from a Gaussian distribution, $N(\mu, \sigma^2\textbf{I}_d)$. The goal is to choose $\mu$ and $\sigma$ carefully so these malicious updates resemble benign ones in their distribution, making them harder to detect. Similarly, the \textit{Noise Attack} involves malicious clients perturbing their local model updates by adding Gaussian noise from $N(\mu, \sigma^2\textbf{I}_d)$. Here, the selection of $\mu$ and $\sigma$ is crucial: too little noise won't affect convergence, while too much will make the malicious updates obviously different from benign ones. Another straightforward technique is the \textit{Sign-flip Attack}, which, instead of altering the magnitude of updates, simply flips the element-wise sign of the model updates to disrupt the aggregation process.}




{\textbf{Advanced Attacks.} Beyond these straightforward, naive approaches, more sophisticated Byzantine attacks employ calculated strategies to subvert FL. The \textit{Min-Max/Min-Sum Attack}~\cite{dnc} is a two-step process: first, attackers perturb the average of benign updates to create a malicious update. Then, for Min-Max, this update is optimized to ensure its maximum Euclidean distance from any benign update doesn't exceed the maximum distance between any two benign updates. In contrast, for Min-Sum, the malicious update is optimized so that the sum of its distances to each benign update is bounded by the maximum total distance of a benign update from others. These attacks are potent because malicious clients strategically search for optimal perturbations to maximize their impact.}

{Another specialized attack is the \textit{AGR-tailored Trimmed-mean Attack}~\cite{dnc}. As proposed by \citet{dnc}, this attack specifically targets defense mechanisms like Trimmed-mean~\cite{trmean} by maximizing the Euclidean distance between the aggregated results of a simple average and the Trimmed-mean defense. While effective and intuitive, the malicious updates generated by this method significantly deviate from benign ones, making them relatively easy for advanced defense mechanisms to detect.}

{In contrast, the \textit{A-little-is-enough (Lie) Attack}~\cite{lie} aims for stealth. Malicious clients apply only a slight perturbation to their local benign updates, making detection difficult. Specifically, they calculate the element-wise mean ($\mu_j$) and standard error ($\sigma_j$) of all updates, then generate elements of malicious updates as $(\beta_i)_j = \mu_j - z \times \sigma_j$, where $\beta$ represents the malicious updates, $j$ is the element index, and $z$ is a scaling factor dependent on the proportion of malicious clients.}

{Finally, the \textit{ByzMean Attack}~\cite{signguard} manipulates the mean of updates to an arbitrarily malicious value. It achieves this by dividing malicious clients into two groups. The first group generates malicious updates using any existing attack method (e.g., Lie attack). The second group then crafts their updates to ensure that the overall average of \textit{all} client updates (benign and malicious) matches the average of the malicious updates from the first group.}

{\textbf{(2) Backdoor Attack in FL.} Backdoor attacks in FL aim to maintain the global model's performance on clean inputs while causing it to make incorrect predictions on inputs containing a specific predefined feature (also called a trigger). Since backdoor attacks~\cite{lyu2024task, lyu2023attention} do not significantly affect the main task accuracy, the malicious local model updates are similar to benign ones, making anomaly detection more challenging in FL. Several classical and commonly discussed backdoor attacks are as follows.}

{\textit{Badnet Attack}~\cite{badnet} is a simple and effective backdoor attack where the attacker injects the same trigger into the training data of compromised clients. The ground truth label of these training data will also be changed to another backdoor class. In contrast to Badnet, the \textit{Distributed Backdoor Attack (DBA)}~\cite{dba} sees the adversary decompose the backdoor trigger into several small triggers and apply them to different compromised clients. This distributed approach makes the attack harder to detect, as each client's updates appear less suspicious. Building upon Badnet, the \textit{Scaling Attack}~\cite{scaling} involves malicious clients scaling their model updates in a way that amplifies the impact of their backdoor-triggered updates without affecting clean data performance. The scaling factor ensures the backdoor pattern has a stronger influence on the global model. The \textit{Projected Gradient Descent (PGD) Attack}~\cite{PGD}, adapted from adversarial machine learning, iteratively perturbs the model updates of compromised clients within a certain bound to embed the backdoor. This method is effective at creating small but impactful changes to the model parameters. Finally, the \textit{Neurotoxin Attack}~\cite{Neurotoxin} explores a persistent attack method designed for scenarios where attackers can only participate in a limited number of rounds. In this attack, malicious clients project their model updates into a subspace that excludes the global coordinates. This approach ensures that the projected updates from malicious clients are primarily embedded in coordinates with minimal perturbation from benign updates, maintaining the backdoor effect even after the attack ceases.}

\begin{wrapfigure}{r}{7.5cm}
\includegraphics[width=7.5cm]{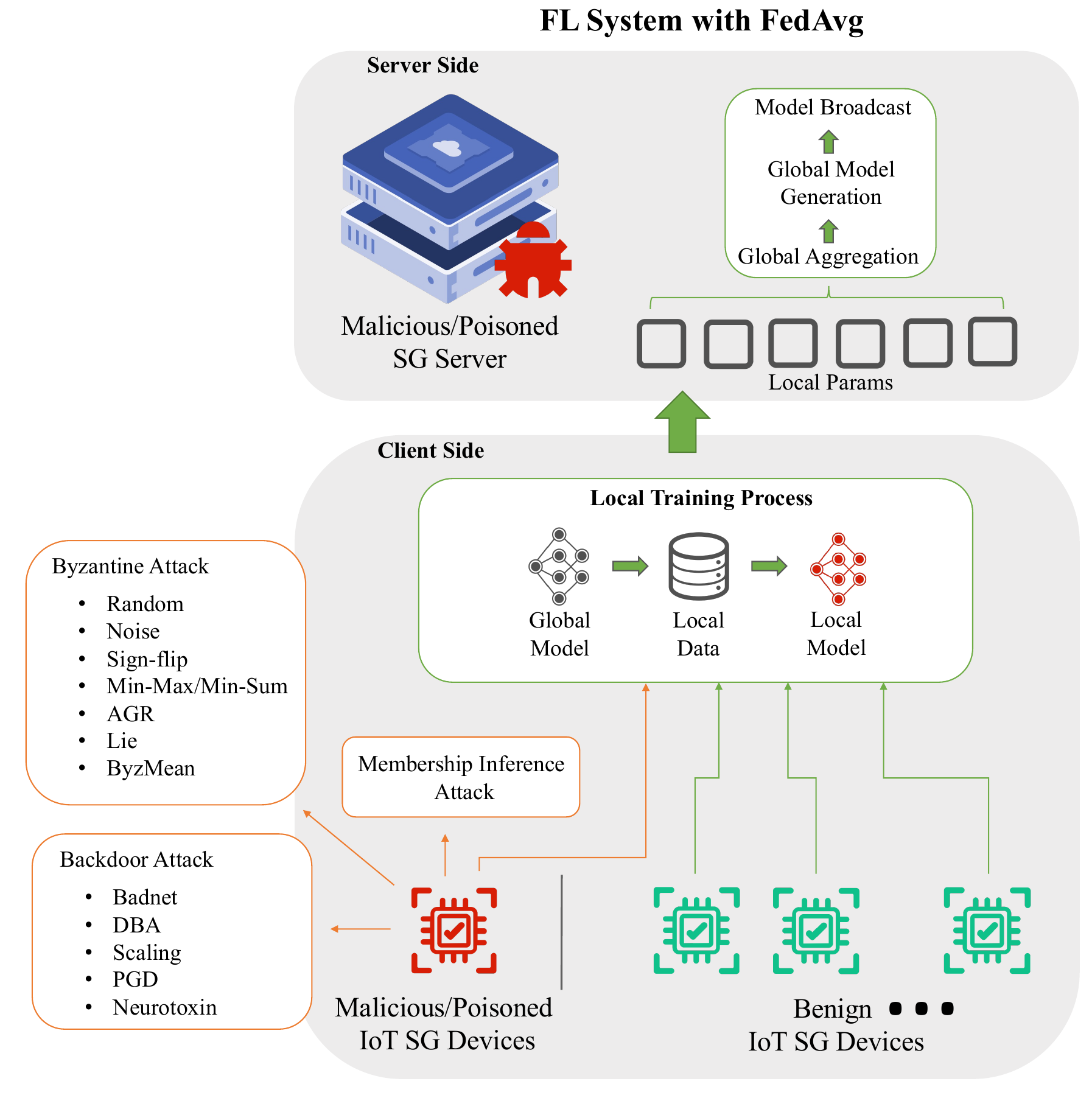}
\caption{The Explanation of the FL Framework and Its Potential Attack Points.\label{fig:fl}}
\end{wrapfigure}

\textbf{(3) Inference Attack in FL.}
Membership Inference Attack (MIA)~\cite{membership1, membership2} is a major privacy attack that seeks to determine whether a specific data record was part of the training dataset used to build a machine learning model. In the context of FL, MIA poses a significant threat to the privacy of participating clients and their data. Various entities, including malicious clients, servers, or external attackers, can execute MIA in FL. These adversaries exploit the model's natural tendency to overfit on training data, leading to unintended information leakage. Approaches to conducting MIA in FL include attack models~\cite{membership3} and gradient-based MIAs, which exploit the gradients exchanged during the FL process. Passive MIA techniques also exist, where adversaries infer membership information by merely observing communication between clients and the server. Successful MIAs compromise the privacy of individual users, a particularly serious issue in sensitive domains such as healthcare or finance, where data confidentiality is paramount. Beyond identifying membership, MIA can also reveal attributes or properties of the training data, further threatening user privacy. Defending against MIA in FL presents unique challenges due to the distributed nature of the learning process, with traditional privacy-preserving techniques either being inapplicable or significantly impacting model utility. As a result, researchers are actively developing novel defense mechanisms to improve FL's resilience to MIA while preserving model performance.

\subsubsection{SOTA Defenses}\label{subsubsec:sota_defenses}
To defend against the attacks mentioned above, the owner of the FL system can adopt various defense methods. In the following, we introduce classical and commonly discussed defense methods. To defend against Byzantine and backdoor attacks, the server can leverage defense methods, which can be generally divided into two main categories: \textit{detection-filtering methods}~\cite{krum, bhagoji2019analyzing, bulyan, LASA, signguard, dnc, fltrust, masa, alignins} and \textit{impact-reduction methods}~\cite{trmean, Foolsgold, lockdown, geomed, sparsefed, zhang2023byzantine}. To defend against inference attacks, one can primarily leverage \textit{differential privacy} (DP)~\cite{ae-dpfl, dp-fedavg, fedsmp, dp-fedsam, gdpfed}.

\textbf{Detection-Filtering Methods.} 
Detection-filtering methods aim to maximize robustness against attacks by identifying and removing malicious local model updates from the aggregation process. These methods typically rely on one or more statistical metrics (e.g., Cosine similarity and Euclidean distance) to assess local model updates. By comparing the extracted metrics with each update, they filter out updates that exhibit abnormal metric values, preventing them from influencing the global model. For example, {Krum}~\cite{krum,bhagoji2019analyzing} selects the most reliable local model that has the smallest sum of squared Euclidean distances to all other models as the output. {Multi-Krum}~\cite{krum} extends this by selecting multiple reliable models for averaging instead of just one. 
\citet{xu2022byzantine} applied the KL-divergence to measure the distance between the empirical sign distributions of the local models in two consecutive rounds and used it to employ a weighted average of models for each client. The DnC proposed in \cite{dnc} calculates the product of a model with its principal component and removes the models with large projection values. {FLTrust}~\cite{fltrust} leverages a root dataset and trains a root global model based on it. After collecting local updates, FLTrust aggregates those updates with high Cosine similarity to root model updates. SignGuard~\cite{signguard} combines direction-based clustering and magnitude-based filtering to identify malicious model updates, where the direction and magnitude are obtained by the sign statistics and $L_2$-norm of the model update, respectively. Recently, a defense method called {LASA} that achieves SOTA results has been proposed in \cite{LASA}. It employs a layer-wise adaptive filter to adaptively select benign sparsified layers using both magnitude and direction metrics across all clients for aggregation.

\textbf{Impact-Reduction Methods.}
Impact-reduction methods try to integrate all model updates but employ strategies to reduce the impact of malicious updates. For instance, {Trimmed mean} (Trmean) proposed in \cite{trmean} discards a certain percentage of the highest and lowest values among the received models from clients for each dimension. After this trimming, the mean of the remaining values is computed by the server, which mitigates the impact of extreme values on the aggregated model. \citet{bulyan} proposed {Bulyan} that combines {Trmean} and {Multi-Krum} to extract benign models. \citet{geomed} proposed to use the {geometric median} (GeoMed) of all local models as the aggregated model, which is effective in filtering out the outlier values. RLR~\cite{rlr} uses the sign information of model updates to adjust the global learning rate for each model coordinate without distinguishing between benign and malicious updates, resulting in poor accuracy.
{Foolsgold}~\cite{Foolsgold} assumes that the malicious updates are consistent with each other. It assigns aggregation weights to model updates based on the maximum Cosine similarity between the last layers of pairwise model updates. A higher Cosine similarity value indicates a higher probability that the updates are malicious, leading to smaller aggregation weights being assigned. {SparseFed} proposed in \cite{sparsefed} sparsifies the aggregated model update at the server side, integrating with model clipping and error feedback, to mitigate the impact of malicious local model updates. In addition, {FedVRDP} proposed in \cite{zhang2023byzantine} applies sparsification on local model updates using a mask obtained from the global model in the previous round and combines it with random Gaussian noise perturbation and variance reduction techniques, to achieve robustness and differential privacy protection simultaneously. A recent approach called {Lockdown}~\cite{lockdown} assumes that those malicious parameters served to recognize backdoor triggers will be deemed unimportant for benign clients and applies sparsification to prune unimportant parameters on the local model.


{\textbf{DP-based Methods.}
The goal of DP is to protect the participation of clients during the training of FL. Specifically, to achieve client-level DP, at training round $t$, each client $i$ clips their local model update $\Delta_i^t$ with a clipping threshold $C$. Subsequently, a small amount of DP noise, drawn from a Gaussian distribution $N(0, C^2\sigma^2 / r \cdot \mathbf{I}_d)$, is applied to the clipped model updates, where $\sigma^2$ is the noise multiplier and $r$ is the number of clients selected to participate in local training per round. When the server receives all perturbed model updates from clients, the accumulative noise becomes $N(0, C^2\sigma^2 \cdot \mathbf{I}_d)$, yielding a rigorous privacy guarantee.}

{However, adding noise to the model updates inherently degrades the model performance of an FL system. Therefore, many methods have been proposed to achieve better performance while maintaining privacy guarantees. For example, the first work to propose client-level DPFL is~\cite{dp-fedavg}, which achieves client-level DP guarantees by employing the Gaussian mechanism and composing privacy guarantees. \citet{blur_lus} explicitly regularize the $L_2$-norm of local updates to ensure they are bounded and further mitigate the negative effects of clipping by applying post-training $\text{Top}_k$ local update sparsification. Fed-SMP~\cite{fedsmp} assumes the server possesses a dataset similar to the overall distribution of client datasets. Using this server dataset, the server selects a set of $\text{Top}_k$ coordinates for all clients to prevent privacy leakage through these selected coordinates. AE-DPFL~\cite{ae-dpfl} ensures user-level DP by employing a voting-based mechanism combined with secure aggregation. Unlike other approaches, AE-DPFL does not require clipping of the model or data, thereby mitigating accuracy degradation. However, AE-DPFL assumes the availability of unlabeled data from the global distribution on the server, which is often impractical in real-world applications. \citet{dp-fedsam} observe that DPFL methods tend to produce sharper loss landscapes, which can make models more sensitive to noise perturbations. To address this issue, they propose DP-FedSAM, which leverages the Sharpness-Aware Minimization (SAM) optimizer. This approach enhances parameter robustness against noise and aims to achieve more stable convergence points. Despite their promise, these methods often assume a uniform privacy requirement across all clients, which is impractical in real-world scenarios.}

\subsection{Discussions and Future Research Directions}\label{subsec:dis_future_rese}
{\textbf{Unique Challenges of FL-based Attack and Defense Methods in SGs.}
Deploying FL in SGs presents unique and substantial challenges that go beyond typical FL implementations, primarily due to the critical nature, scale, and inherent characteristics of these energy infrastructures. Successfully bridging the gap between SOTA FL methodologies and their practical, secure application in SGs requires addressing these multifaceted issues.}

{One of the foremost challenges stems from the resource-constrained nature of many SG devices. Components like smart meters, sensors, and actuators often possess limited computational power, memory, and energy budgets. This directly impacts the complexity of local model training permissible within an FL framework and constrains the sophistication of on-device operations. From an attack perspective, while simple disruptive actions like naive Byzantine attacks (e.g., sign-flips or random noise injection) might be feasible from such constrained devices, more advanced, computationally intensive attacks (e.g., optimization-based Byzantine attacks or complex local data poisoning for backdoors) might be limited to more powerful edge devices or orchestrated by a well-resourced adversary controlling multiple simpler nodes. Similarly, the implementation of robust on-device defenses or complex local differential privacy mechanisms can be severely hampered, forcing reliance on server-side or hierarchical defense strategies.}

{The integration of FL with existing SG infrastructure and its real-time operational demands poses another significant hurdle. SGs are often a hybrid of legacy systems and modern digital components, leading to interoperability issues, diverse communication protocols (each with potential vulnerabilities), and difficulties in retrofitting new technologies like FL securely. The real-time requirements for many SG applications, such as state estimation, fault detection, or dynamic load balancing, mean that FL models must be highly accurate and responsive. This critical dependency makes the SG system particularly vulnerable to attacks that degrade model performance or introduce latency. For example, Byzantine attacks that corrupt model updates or backdoor attacks that cause misbehavior on specific trigger conditions can have severe physical consequences, from economic losses to power outages. Consequently, defense mechanisms must not only be effective but also computationally efficient to operate within these low-latency constraints, and their false positive rates must be exceptionally low to prevent disruption of essential grid services.}

{Furthermore, SG data characteristics and governance complexities uniquely challenge FL deployment. SG data is inherently heterogeneous and non-IID due to varying consumer behaviors, geographical factors, and diverse energy resources. This non-IID nature can impede the convergence of global FL models, introduce biases, and significantly complicate the task of server-side defenses. Detection-filtering methods (e.g., Krum, statistical outlier removal) may struggle to distinguish between genuinely malicious updates and those that are merely statistically different due to legitimate local data variations. The sensitive nature of energy consumption data also makes SGs a prime target for inference attacks, such as Membership Inference Attacks, where adversaries attempt to glean private information from model updates or outputs. While FL aims to protect raw data, the model itself becomes a conduit for potential leakage. Implementing defenses like DP is crucial, but the inherent trade-off between privacy and model utility is particularly acute in SGs; the noise introduced to ensure privacy can degrade the accuracy of critical control and forecasting functions, demanding meticulous calibration.}

{Finally, the overall attack surface and the practical feasibility of robust defenses in a distributed SG environment present an overarching challenge. While FL decentralizes data, it introduces numerous potential points of failure or compromise through its participating clients (SG devices). Implementing sophisticated, coordinated attacks (e.g., ByzMean, distributed backdoors) becomes a question of an attacker's ability to compromise a sufficient number of these distributed entities. On the defense side, there's a constant tension between the need for robust, computationally intensive aggregation-stage defenses at the server (or edge aggregators) and the communication overhead and trust assumptions associated with collecting necessary information (like update similarities or auxiliary data) from potentially untrustworthy participants. The selection and tuning of defenses—whether detection-based, impact-reduction, or privacy-enhancing—must be carefully tailored to the specific SG application, the capabilities of participating devices, and the acceptable risk tolerance for critical infrastructure.}

{\textbf{Future Directions for FL in Smart Grids.} To advance the practical and secure deployment of FL in SG systems, future research must strategically address fundamental security, privacy, and operational challenges. A primary focus should be the development of highly tailored security and privacy frameworks robust enough for critical SG infrastructure. This involves creating adaptive, resource-aware defense mechanisms capable of mitigating sophisticated attacks (e.g., Byzantine, backdoor, inference) within heterogeneous SG environments characterized by resource-constrained devices and non-IID data. Concurrently, significant advancements are required in utility-preserving privacy-enhancing technologies, such as refined DP schemes or practical hybrid cryptographic solutions, to protect sensitive SG data during critical applications like state estimation or load forecasting, without unduly compromising model accuracy or operational efficiency, while also ensuring verifiable compliance with regulatory standards.}

{Furthermore, enhancing the operational viability and real-world applicability of FL in dynamic SG settings is paramount. A key direction is advancing asynchronous FL (AFL) protocols, which are inherently better suited to the SG's typical device heterogeneity, variable network conditions, and potential for intermittent client connectivity than traditional synchronous approaches. Building upon existing research into Byzantine-resilient asynchronous methods~\cite{asy1,asy2,asy3}, future work should develop lightweight yet robust AFL aggregation techniques specifically for SG data characteristics. This includes analyzing the distinct impact of asynchronicity on various attack surfaces and defense dynamics, ensuring model convergence and fairness under realistic SG conditions, and designing effective incentive mechanisms. Successfully tailoring and securing AFL for SGs will significantly improve system efficiency, scalability, and overall resilience in complex, large-scale deployments.}

\section{{\textit{FedGridShield}: An Open-Source Framework for FL-based Attack and Defense Methods in Smart Grids}}\label{sec:fedgridshield}

{We introduce \textit{FedGridShield}\footnote{https://github.com/FedGridShield/FedGridShield}, a lightweight and scalable framework designed to implement FL-based attack and defense methods in SGs. FedGridShield supports ongoing cybersecurity research in the interdisciplinary field of FL and SGs by providing a user-friendly platform capable of handling various attack and defense methods and SG datasets.}

{The current framework includes attack methods such as AGR-tailored Trimmed-mean Attack~\cite{dnc} and Fang Attack~\cite{fang2020local}; {defense} methods including Trimmed Mean~\cite{trmean}, Median~\cite{trmean_median}, Krum~\cite{krum}, Bulyan~\cite{bulyan}, and FLAME~\cite{nguyen2022flame}; and {privacy-preserving} methods DPFed~\cite{zhang2023byzantine} and SparseFed~\cite{sparsefed}.}

{We conduct experiments on two datasets: {Electricity Theft Detection:} We use data introduced by \citet{zhao2023privacy}, which contains the power consumption data of 200 consumers, resulting in a total of 107,200 records. This is a tabular dataset with 48 features, and the binary label defines whether electricity theft is detected. {Generator Defect Classification:} We also collect a dataset of generator images in perfect and defective conditions. This dataset is designed to train an image classifier to indicate the generator's condition, which can help with system maintenance. The training data contains 766 images of generators in perfect condition and 744 images of generators in defective condition. The test data contains 275 and 300 images for these conditions, respectively. The results are reported on the project page due to page {restrictions}.}

\section{Conclusions}\label{conclusion}

In this paper, we present a comprehensive review of studies focusing on FL-based applications in SGs and their potential vulnerabilities. We begin by comparing our work with existing surveys and outlining our unique contributions. 
{Following this, we define key terminologies related to SGs and FL, outline common FL threat models, and provide an overview of frequently discussed attack methods within FL frameworks. Next, we examine the deployment of FL across the three key stages of SG systems: generation, transmission and distribution, and consumption. The integration of FL at these stages has been shown to significantly enhance data privacy and distribution efficiency in SGs. Furthermore, we analyze the potential vulnerabilities inherent in these FL-based SG systems. We also introduce the FedGridShield framework, designed to facilitate further research and help address identified gaps. Finally, this paper summarizes existing research on FL vulnerabilities and proposes key directions for future investigation.}

\begin{acks}
The work of S. Rath was supported by the Faculty Startup Grant at the University of Tulsa.
\end{acks}

\bibliographystyle{ACM-Reference-Format}
\bibliography{main}

\end{document}